\documentclass[runningheads]{llncs}
\usepackage{graphicx}
\usepackage{subcaption}
\usepackage{hyperref}
\usepackage{xspace}
\usepackage{algorithm}
\usepackage{endnotes}
\usepackage{etoolbox}
\usepackage{algpseudocode}
\usepackage{amsmath}
\usepackage{bm}

\newcommand{\model}{GIPA\xspace}
\begin{document}
\title{GIPA: A General Information Propagation Algorithm for Graph Learning}

%
%
\author{
Houyi Li\inst{1}\thanks{Contributed to this work when the author worked in Ant and Alibaba Group.} \and
Zhihong Chen\inst{2} \and
Zhao Li\inst{3, 6} \and
Qinkai Zheng \inst{4} \and
Peng Zhang \inst{5}\and
Shuigeng Zhou\inst{1}\thanks{Shuigeng Zhou is the corresponding author.}}
\authorrunning{Li, H. et al.}
%
\institute{School of computer science, Fudan University, Shanghai, China \and
Alibaba Group, Hangzhou, China \and
Zhejiang University, Hangzhou, China \and 
Tsinghua University, Beijing, China \and 
Cyberspace Institute of Advanced Technology, Guangzhou University, China \and 
Link2Do Technology, Hangzhou, China\\
\email{lihouyi2008@126.com, jhon.czh@alibaba-inc.com, lzjoey@gmail.com, qinkai.zheng1028@gmail.com, p.zhang@gzhu.edu.cn, sgzhou@fudan.edu.cn}
}
\maketitle              
\begin{abstract}
Graph neural networks (GNNs) have been widely used in graph-structured data computation, showing promising performance in various applications such as node classification, link prediction, and network recommendation.
Existing works mainly focus on node-wise correlation when doing weighted aggregation of neighboring nodes based on attention, such as dot product by the dense vectors of two nodes. This may cause conflicting noise in nodes to be propagated when doing information propagation. To solve this problem, we propose a General Information Propagation Algorithm (\model), which exploits more fine-grained information fusion including bit-wise and feature-wise correlations based on edge features in their propagation. Specifically, the bit-wise correlation calculates the element-wise attention weights through a multi-layer perceptron (MLP) based on the dense representations of two nodes and their edge; The feature-wise correlation is based on the one-hot representations of node attribute features for feature selection. We evaluate the performance of \model on the Open Graph Benchmark proteins (OGBN-proteins) dataset and the Alipay dataset of Alibaba Group.
Experimental results reveal that \model outperforms the state-of-the-art models in terms of prediction accuracy, e.g., \model achieves an average ROC-AUC of $0.8917\pm 0.0007$, which is better than that of all the existing methods listed in the OGBN-proteins leaderboard.

\keywords{Graph neural networks, Fine-grained information fusion,  Bit-wise and feature-wise attention.}
\end{abstract}
\section{Introduction}
Graph representation learning typically aims to learn an informative embedding for each graph node based on the graph topology (link) information.
Generally, the embedding of a node is represented as a low-dimensional feature vector, which can be used to facilitate downstream applications.
This research focuses on homogeneous graphs that have only one type of nodes and one type of edges. The purpose is to learn node representations from the graph topology~\cite{grover2016node2vec,perozzi2014deepwalk,dai2016discriminative}.
Specifically, given a node $u$, either breadth-first search, depth-first search or random walk is used to identify a set of neighboring nodes. Then, $u$'s embedding is learnt by maximizing the co-occurrence probability of $u$ and its neighbors.
Early studies on graph embedding have limited capability to capture neighboring information from a graph because they are based on shallow learning models such as SkipGram~\cite{mikolov2013distributed}.
Moreover, transductive learning is used in these graph embedding methods, which cannot be generalized to new nodes that are absent in the training graph.

Graph neural networks~\cite{kipf2016semi,hamilton2017inductive,velivckovic2017graph} are proposed to overcome the limitations of traditional graph embedding models.
GNNs employ deep neural networks to aggregate feature information from neighboring nodes and thereby have the potential to gain better aggregated embeddings.
GNNs can support inductive learning and infer the class labels of unseen nodes during prediction~\cite{hamilton2017inductive,velivckovic2017graph}.
The success of GNNs is mainly due to the neighborhood information aggregation. However, 
GNNs face two challenges: \textit{which neighboring nodes of a target node are involved in message passing? and how much contribution each neighboring node makes to the aggregated embedding?}.
For the former question, neighborhood sampling~\cite{hamilton2017inductive,ying2018graph,chen2018fastgcn,huang2018adaptive,zou2019layer,ji2020accelerating} is proposed for large dense or power-law graphs.
For the latter, neighbor importance estimation is used to attach different weights to different neighboring nodes during feature propagation.
Importance sampling~\cite{chen2018fastgcn,zou2019layer,ji2020accelerating} and attention~\cite{velivckovic2017graph,liu2019geniepath,wang2019heterogeneous,yun2019graph,hu2020heterogeneous} are two popular techniques.
Importance sampling is a special case of neighborhood sampling, where the importance weight of a neighboring node is drawn from a distribution over nodes.
This distribution can be derived from normalized Laplacian matrices~\cite{chen2018fastgcn,zou2019layer} or jointly learned with GNNs~\cite{ji2020accelerating}.
With this distribution, at each step a subset of neighbors is sampled, and aggregated with the importance weights.
Similar to importance sampling, attention also attaches importance weights to neighbors.
Nevertheless, attention differs from importance sampling.
Attention is represented as a neural network and is always learned as a part of a GNN model.
In contrast, importance sampling algorithms use statistical models without trainable parameters.

Existing attention mechanisms consider only the correlation of node-wise, ignoring the suppression of noise information in transmission, and the information of edge features. In real world applications, only partial users authorize the system to collect theirs profiles. The model cannot learn the node-wise correlation between a profiled user and a user we know nothing about.  Therefore, existing models will spread noise information, resulting in inaccurate node representations. However, two users who often transfer money to each other and two users who only have a few conversations have different correlation. 


In this paper, to solve the problems mentioned above, we present a new graph neural network attention model, namely \underline{G}eneral \underline{I}nformation \underline{P}ropagation \underline{A}lgorithm (\model). We design a bit-wise correlation module and a feature-wise correlation module. Specifically, we believe that each dimension of the dense vector represents a feature of the node. Therefore, the bit-wise correlation module filters at the dense representation level. The dimension of attention weights is equal to that of density vector. In addition, we represent each attribute feature of the node as a one-hot vector. The feature-wise correlation module performs feature selection by outputting the attention weights of similar dimensionality and attribute features. It is worth mentioning that to enable the model to extract better attention weights, edge features that measure the correlation between nodes are also included in the calculation of attention. Finally, \model inputs sparse embedding and dense embedding into the wide end and deep end of the deep neural network for learning specific tasks, respectively. 
Our contributions are summarized as follows:
\begin{itemize}
  \item[1)]
We design the bit-wise correlation module and the feature-wise correlation module to perform more refined information weighted aggregation from the element level and the feature level, and utilize the edge information.
  \item[2)]
 Based on the wide \& deep architecture~\cite{WideDeep}, we use dense feature representation and sparse feature representation to extract deep information and retain shallow original information respectively, which provides more comprehensive information for the downstream tasks. 
 \item[3)]
 Experiments on the Open Graph Benchmark (OGB)~\cite{hu2020open} proteins dataset (OGBN-proteins) demonstrate that \model achieves better accuracy with an average ROC-AUC of $0.8917\pm 0.0007$~\footnote{The reproducible code is open source: https://github.com/houyili/gipa\_wide\_deep} than the state-of-the-art methods listed in the OGBN-proteins leaderboard~\footnote{https://ogb.stanford.edu/docs/leader\_nodeprop/\#ogbn-proteins}.
 In addition, \model has been tested on billion-scale industrial Alipay dataset.
 
\end{itemize}

\section{Related Work}
In this section, we review existing attention primitive implementations in brief.
\cite{bahdanau2014neural} proposes an additive attention that calculates the attention alignment score using a simple feed-forward neural network with only one hidden layer.
The alignment score $score(q,k)$ between two vectors $q$ and $k$ is defined as
$score(q,k) = u^T\tanh(W[q;k])$, 
where $u$ is an attention vector and the attention weight $\alpha_{q,k}$ is computed by normalizing $scor\tilde{e}_{q,k}$ over all ${q,k}$ values with the softmax function.
The core of the additive attention lies in the use of attention vector $u$.
This idea has been widely adopted by several algorithms~\cite{yang2016hierarchical,pavlopoulos2017deeper} for natural language processing.
\cite{luong2015effective} introduces a global attention and a local attention.
In global attention, the alignment score can be computed by three alternatives: dot-product ($q^Tk$), general ($q^TWk$) and concat ($W[q;k]$).
In contrast, local attention computes the alignment score solely from a vector ($Wq$).
Likewise, both global and local attention normalize the alignment scores with the \textit{softmax} function.
\cite{vaswani2017attention} proposes a self-attention mechanism based on scaled dot-product.
This self-attention computes the alignment score between any $q$ and $k$ as follows: 
$scor\tilde{e}_{q,k} = {q^Tk}\big/ {\sqrt{d_k}}$. 
%
This attention differs from the dot-product attention~\cite{luong2015effective} by only a scaling factor of $\frac{1}{\sqrt{d_k}}$.
The scaling factor is used because the authors of~\cite{vaswani2017attention} suspected that for large values of $d_k$, dot-product is large in magnitude, which thereby pushes the \textit{softmax} function into regions where it has extremely small gradients.
More attention mechanisms, like feature-wise attention~\cite{FeatureAtt} and multi-head attention~\cite{vaswani2017attention}, can be referred to some surveys~\cite{attsurvey,NIU202148}.

In addition, these aforementioned attention primitives have been extended to heterogeneous graphs.
HAN~\cite{wang2019heterogeneous} uses a two-level hierarchical attention that consists of a node-level attention and a semantic-level attention.
In HAN, the node-level attention learns the importance of a node to any other node in a meta-path, while the semantic-level one weighs all meta-paths.
HGT~\cite{zhang2019heterogeneous} weakens the dependency on meta-paths and instead uses meta-relation triplets as basic units.
HGT uses node-type-aware feature transformation functions and edge-type-aware multi-head attention to compute the importance of each edge to a target node.
It is worthy of mentioning that heterogeneous models must not always be superior to homogeneous ones, and vice versa. In this paper, unlike the node-wise attention in existing methods that ignores noise propagation, our proposed \model introduces two MLP-based correlation modules, the bit-wise correlation module and the feature-wise correlation module, to achieve fine-grained selective information propagation and utilize the edge information.

\section{Methodology}
\subsection{Preliminaries}
\paragraph{Graph Neural Networks.} Consider an attributed graph $\mathcal{G}=\{\mathcal{V}, \mathcal{E}\}$, where $\mathcal{V}$ is the set of nodes and $\mathcal{E}$ is the set of edges. GNNs use the same model framework as follows~\cite{hamilton2017inductive,xu2018powerful}:

\begin{figure*}[t!]
    \centering
    \includegraphics[width=0.95\textwidth]{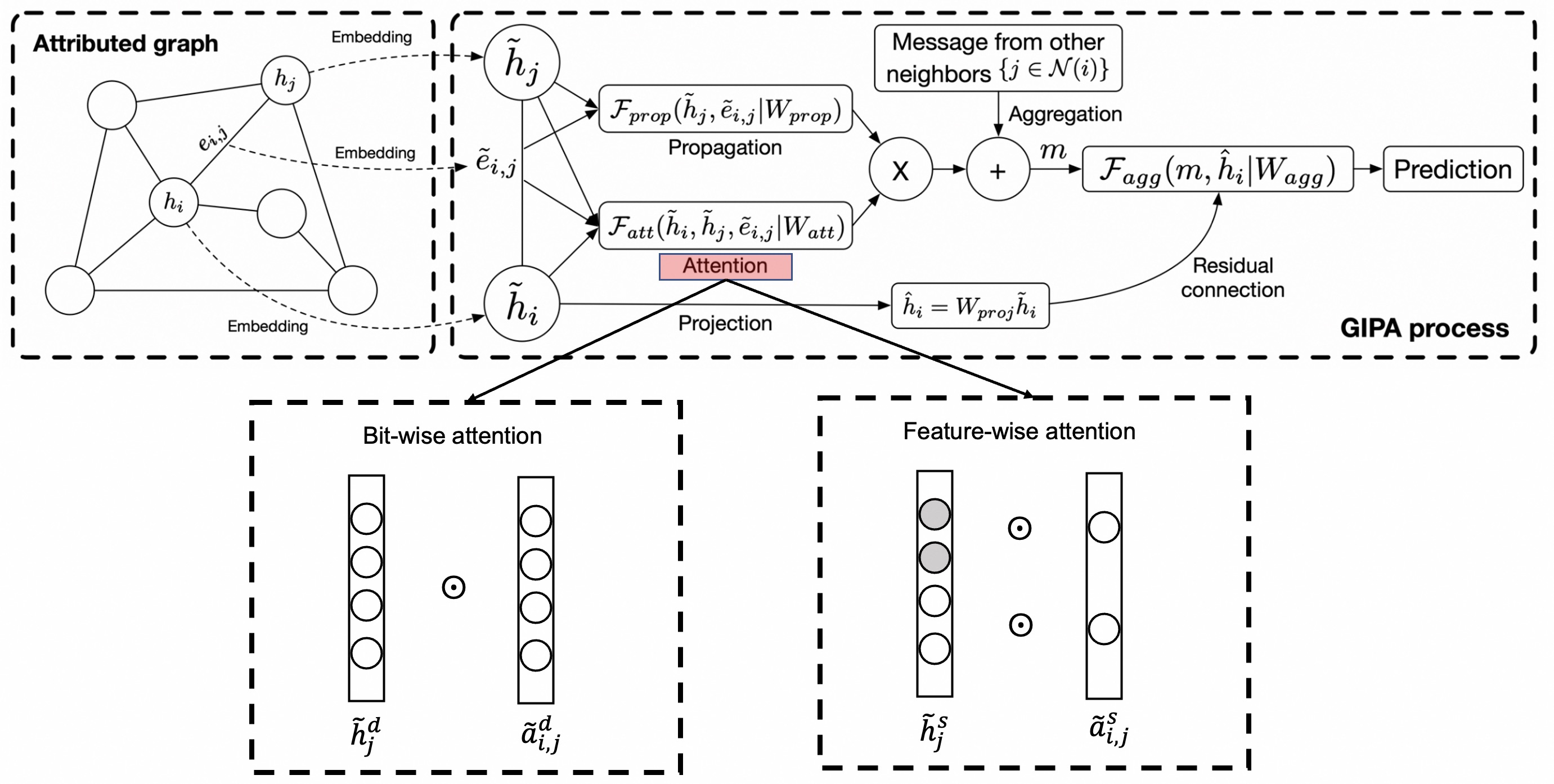}
    \caption{The architecture of \model, which consists of \textit{attention}, \textit{propagation}, and \textit{aggregation} modules (or processes). The red shadow indicates the bit-wise module and feature-wise module, which extract neighbor information more relevant to the current node through more fine-grained selective information fusion.}
    \label{fig:gipa_process}
\vspace{-0.1cm}
\end{figure*}

\begin{equation}
\label{eq:gnn}
\tilde{h}_i^k=\phi(\tilde{h}_i^{k-1}, \mathcal{F}_{agg}(\{\tilde{h}_j^{k-1}, j\in\mathcal{N}(i)\}))
\end{equation}
and \noindent where $\mathcal{F}_{agg}$ represents an aggregation function, $\phi$ represents an update function, $\tilde{h}_i^k$ represents the node embedding based on node feature $x_i^k$, and $\mathcal{N}(i)$ represents the neighbor node set. The objective of GNNs is to update the embedding of each node by aggregating the information from its neighbor nodes and the connections between them. 

\subsection{Model Architecture}
In this section, we present the architecture of \model in~\figurename~\ref{fig:gipa_process}, which  extracts node features and edge features in a more general way. Consider a node $i$ with feature embedding $\tilde{h}_i$ and its neighbor nodes $j\in\mathcal{N}(i)$ with feature embedding $\tilde{h}_j$. $\tilde{e}_{i, j}$ represents the edge feature between node $i$ and $j$. The problem is how to generate an expected embedding for node $i$ from its own node feature embedding $\tilde{h}_i$, its neighbors' node features embedding $\{\tilde{h}_j\}$ and the related edge features $\{\tilde{e}_{i, j}\}$. 
The workflow of \model consists of three major modules (or processes): \textit{attention}, \textit{propagation} and \textit{aggregation}. 
First, \model computes the dense embedding $\tilde{h}_i^d$, $\tilde{h}_j^d$, and $\tilde{e}_{i, j}$, and sparse embedding $\tilde{h}_i^s$, $\tilde{h}_j^s$ from raw features.
Then, the \textit{attention} process calculates the bit-wise attention weights by using fully-connected layers on $\tilde{h}_i^d$, $\tilde{h}_j^d$, and $\tilde{e}_{i, j}$ and feature-wise by using fully-connected layers on $\tilde{h}_i^s$, $\tilde{h}_j^s$, and $\tilde{e}_{i, j}$. 
Following that, the \textit{propagation} process focuses on propagating information of each neighbor node $j$ by combining $j$'s node embedding $\tilde{h}_j$ with the associated edge embedding $\tilde{e}_{i, j}$. 
Finally, the \textit{aggregation} process aggregates all messages from neighbors to update the embedding of $i$. The following subsections introduce the details of each process in one layer.

\subsubsection{Embedding Layer.}
The \model (wide \& deep) is more suitable for these scenarios: each node in GNN is not composed of text or images, but represents objects such as users, products, and proteins, whose features are composed of category features and statistical features. 
For example, the category feature in \textit{ogbn-proteins} dataset is what species the proteins come from. 
And the statistical features in \textit{Alipay} dataset are similar to the total consumption of users in a year.

For dense embedding, each integer number can express a category, each floating-point number can express a statistical value, thus a one-dimensional embedding can represent one feature.
However, the sparse embedding of one feature requires more dimensions.
For each category feature, one-hot encoding is required. 
For example, a certain ``category feature" has a total of $K$ possible categories, and an $K+1$ dimensional vector is required to represent the feature.
And each "statistical feature" is cut into $K$ categories (using equal-frequency or equal-width cutting method), and then $K+1$ dimensional one-hot encoding is performed.
Thus, the concatenations of dense and sparse embeddings are the inputs of deep part and wide part respectively.

\begin{algorithm}[t]
\caption{The optimization strategy of GIPA}
\label{alg::algorithm1}
\begin{algorithmic}[1]
\State \textbf{Input:} Graph $\mathcal{G}=\{\mathcal{V}, \mathcal{E}\}$; input features $\{x_{v}, \forall v \in V\}$; Number of layer $K$
\State \textbf{Note:} $\tilde{h}_{v}^* \in \{\tilde{h}_{v}^s,   \tilde{h}_{v}^d\}$
\State $\tilde{h}_v^{d^0} \leftarrow DenseEmb(x_{v}), \forall v \in V$  \State $\tilde{h}_v^{s^0} \leftarrow SparseEmb(x_{v}), \forall v \in V$
\State\textbf{while} not end of epoch \textbf{do}
\State \qquad Select a subgraph $\mathcal{G}_t=\{\mathcal{V}_t, \mathcal{E}_t\} \in \mathcal{G}$
\State \qquad \textbf{for} each $k \in [1, K]$ \textbf{do}
\State \qquad \qquad \textbf{if} $k > 1$
\State \qquad \qquad \qquad $\tilde{h}_i^{*k-1} \leftarrow o_{i}^{*k-1}, \forall v_i \in V_t$
\State \qquad \qquad $\tilde{\bm{a}}_{i, j}^{*k} \leftarrow \mathcal{F}_{act}(\mathcal{F}_{att}^*(\tilde{h}_i^{*k-1}, \tilde{h}_j^{*k-1}, \tilde{e}_{i, j}|W_{att}^*)), \forall v_i \in V_t$
\State \qquad \qquad $m_{i, j}^{*k} \leftarrow \tilde{\bm{a}}_{i, j}^{*k} * \mathcal{F}_{prop}^*(\tilde{h}_j^{*k-1}, \tilde{e}_{i,j}|W_{prop}^*), \forall v_i \in V_t$
\State \qquad \qquad $o_{i}^{*k} \leftarrow \mathcal{F}_{agg}^*(\sum_{j\in\mathcal{N}(i)}m_{i, j}^{*k}, \hat{h}_i^{*k-1} | W_{agg}^*), \forall v_i \in V_t$
\State \qquad $\hat{y}_i = Deep(o_{i}^{dk})+ W_{wide}(o_{i}^{sk})$
\State \qquad $L \leftarrow \sum_{\forall v_i \in V_t}l(y_i, \hat{y}_i)$
\State \qquad Update all parameters by gradient of $L$
\end{algorithmic}
\end{algorithm}

\subsubsection{Attention Process.}
Different from the existing attention mechanisms like self-attention or scaled dot-product attention, we use MLP to realize a bit-wise attention mechanism and a feature-wise attention mechanism.
The bit-wise  and feature-wise \textit{attention} process of \model can be formulated as follows:

\begin{equation}
\label{eq:att1}
\bm{a}_{i, j}^d = \mathcal{F}_{att}^d(\tilde{h}_i^d, \tilde{h}_j^d, \tilde{e}_{i, j}|W_{att}^d) = \text{MLP}([\tilde{h}_i^d || \tilde{h}_j^d || \tilde{e}_{i, j}]|W_{att}^d)
\end{equation}

\begin{equation}
\label{eq:att2}
\bm{a}_{i, j}^s = \mathcal{F}_{att}^s(\tilde{h}_i^s, \tilde{h}_j^s, \tilde{e}_{i, j}|W_{att}^s) = \text{MLP}([\tilde{h}_i^s || \tilde{h}_j^s || \tilde{e}_{i, j}]|W_{att}^s)
\end{equation}
\noindent where $\bm{a}_{i, j}^d \in \mathcal{R}^n$ is bit-wise attention weight and its dimension is the same as that of $\tilde{h}_i^d$, $\bm{a}_{i, j}^s \in \mathcal{R}^m$ is feature-wise attention weight and its dimension is the same as the number of node features, the attention function $\mathcal{F}_{att}^*$ is realized by an \textit{MLP} with learnable weights $W_{att}^*$ (without bias). Its input is the concatenation of the node embeddings $\tilde{h}_i^*$ and $\tilde{h}_j^*$ as well as the edge embedding $\tilde{e}_{i, j}$. As the edge features $\tilde{e}_{i, j}$ measuring the correlation between nodes are input into \textit{MLP}, this attention mechanism could be more representative than previous ones simply based on dot-product. The final attention weight is calculated by an  activation function for bit-wise part and feature-wise part:
\begin{equation}
\label{eq:softmax1}
\tilde{\bm{a}}_{i, j}^* = \mathcal{F}_{act}(\bm{a}_{i, j}^*) 
\end{equation}
%
\noindent where $ \mathcal{F}_{act}$ represents the activation function, such as \textit{softmax}, \textit{leaky-relu}, \textit{softplus},  etc. Based on the experimental results, we finally define the activation function as \textit{softplus}.  Details can be seen in Section \ref{Hyperparameter}.

\subsubsection{Propagation Process.}
Unlike GAT~\cite{velivckovic2017graph} that considers only the node feature of neighbors, \model incorporates both node and edge embeddings during the \textit{propagation} process:
\begin{equation}
\label{eq:prop1}
p_{i, j}^d = \mathcal{F}_{prop}^d(\tilde{h}_j^d, \tilde{e}_{i,j}|W_{prop}^d) = \text{MLP}([\tilde{h}_j^d || \tilde{e}_{i, j}]|W_{prop}^d), 
\end{equation}

\begin{equation}
\label{eq:prop1}
p_{i, j}^s = \mathcal{F}_{prop}^s(\tilde{h}_j^s, \tilde{e}_{i,j}|W_{prop}^s) = \text{MLP}([\tilde{h}_j^s || \tilde{e}_{i, j}]|W_{prop}^s), 
\end{equation}
\noindent where the propagation function $\mathcal{F}_{prop}^*$ is also realized by an \text{MLP} with learnable weights $W_{prop}^*$. Its input is the concatenation of a neighbor node dense and sparse embeddings $\tilde{h}_j^*$ and the related edge embedding $\tilde{e}_{i, j}$. Thus, the $propagation$ is done bit-wise and feature-wise rather than node-wise. 

Combining the results by \textit{attention} and \textit{propagation} by bit-wise and feature-wise multiplication, \model gets the message $m_{i, j}^d$ and $m_{i, j}^s$ of node $i$ from $j$:
\begin{equation}
\label{eq:msg}
m_{i, j}^d = \tilde{\bm{a}}_{i, j}^d * p_{i, j}^d \ \ \ m_{i, j}^s = \tilde{\bm{a}}_{i, j}^s \otimes p_{i, j}^s
\end{equation}


\subsubsection{Aggregation Process.}
For each node $i$, \model repeats previous processes to get messages from its neighbors. The \textit{aggregation} process first gathers all these messages by a reduce function, summation for example:

\begin{equation}
\label{eq:sum}
m_{i}^{*} = \sum_{j\in\mathcal{N}(i)}m_{i, j}^{*}
\end{equation}


\noindent Then, a residual connection between the linear projection $\hat{h}_i^*$ and the message of $m_i$ is added through concatenation: 



\begin{equation}
\label{eq:proj}
\hat{h}_i^d=W_{proj}^d\tilde{h}_i^d \ \ \ \hat{h}_i^s=W_{proj}^s\tilde{h}_i^s
\end{equation}

\begin{equation}
\label{eq:agg}
o_{i}^d = \mathcal{F}_{agg}^d(m_{i}^d, \tilde{h}_i^d | W_{agg}^d) = \text{MLP}([m_{i}^d || \tilde{h}_i^d]|W_{agg}^d) \oplus \hat{h}_i^d
\end{equation}

\begin{equation}
\label{eq:agg}
o_{i}^s = \mathcal{F}_{agg}^s(m_{i}^s, \tilde{h}_i^s | W_{agg}^s) = \text{MLP}([m_{i}^s || \tilde{h}_i^s]|W_{agg}^s) \oplus \hat{h}_i^s
\end{equation}

\noindent where an \textit{MLP} with learnable weights $W_{agg}^*$ is applied to get the final dense output $o_{i}^d$ and sparse output $o_{i}^s$ . 
Finally, we would like to emphasize that the process of \model can be easily extended to multi-layer variants by stacking the process multiple times. After we get the aggregated output of the node, $o_{i}^d$ and $o_{i}^s$ respectively input the depth side and wide side of the Deep\&Wide architecture for downstream tasks. See algorithm \ref{alg::algorithm1} for details.

\section{Experiments}
\subsection{Datasets and Settings}
\noindent \textbf{Datasets.} In our experiments, we choose two edge-attribute dataset: the \textit{ogbn-proteins} dataset from OGB~\cite{hu2020open} and \textit{Alipay} dataset\cite{graphtheta}. The \textit{ogbn-proteins} dataset is an undirected and weighted graph, containing 132,534 nodes of 8 different species and 79,122,504 edges with 8-dimensional features. The task is a multi-label binary classification problem with 112 classes representing different protein functions. The \textit{Alipay} dataset is an edge attributed graph, containing 1.40 billion nodes with 575 features and 4.14 billion edges with 57 features. The task is a multi-label binary classification problem. It is worth noting that due to the 
high cost of training on \textit{Alipay} data set, we only conduct ablation experiments on the
input features of $\mathcal{F}_{att}$ and $\mathcal{F}_{prop}$ in the industrial data set, as Table \ref{tab:ali_performance}.

\noindent \textbf{Baselines.} Several representative GNNs including SOTA GNNs are used as baselines. For semi-supervised node classification, we utilize GCN \cite{kipf2016semi}, GraphSAGE \cite{hamilton2017inductive}, GAT \cite{velivckovic2017graph}, MixHop \cite{abu2019mixhop}, JKNet \cite{xu2018representation}, DeeperGCN \cite{li2020deepergcn}, GCNII \cite{chen2020simple}, DAGNN \cite{liu2020towards}, MAGNA \cite{wang2020direct}, UniMP \cite{shi2020masked}, GAT+BoT \cite{wang2021bag}, RevGNN \cite{li2021training} and AGDN \cite{sun2020adaptive}. Note that DeeperGCN, UniMP, RevGNN, and AGDN are implemented with random partition. GAT is implemented with neighbor sampling. Except for our \model, results of other methods are from their papers or the OGB leaderboard.

\noindent \textbf{Evaluation metric.} The performance is measured by the average ROC-AUC scores. We follow the dataset splitting settings as recommended in OGB and test the performance of 10 different trained models with different random seeds. 

\noindent \textbf{Hyperparameters.} For the number of layers $K$, we search the best value from 1 to 6. As for the activation function of attention process, we consider common activation functions. For details, please refer to Section~\ref{Hyperparameter}.

\noindent \textbf{Running environment.} For \textit{ogbn-proteins} dataset, \model is implemented in Deep Graph Library (DGL)~\cite{wang2019deep} with Pytorch~\cite{paszke2019pytorch} as the backend. Experiments are done in a platform with Tesla V100 (32G RAM). For \textit{Alipay} dataset, \model is implemented in \textit{GraphTheta}~\cite{graphtheta}, and runs on private cloud of Alibaba Group.

\begin{table}[!h]
\centering
\caption{Test and validation performance results (ROC-AUC) on the \textit{ogbn-proteins} dataset. The improvements over comparison methods are statistically significant at 0.05 level.}
\begin{tabular}{lcc}
\hline
Method              & Test ROC-AUC  & Validation ROC-AUC \\ \hline
GCN                 & $0.7251\pm0.0035$ & $0.7921\pm0.0018$      \\
GraphSAGE           & $0.7768\pm0.0020$ & $0.8334\pm0.0013$      \\
DeeperGCN           & $0.8580\pm0.0028$ & $0.9106\pm0.0011$      \\
GAT                 & $0.8682\pm0.0018$ & $0.9194\pm0.0003$      \\
UniMP               & $0.8642\pm0.0008$ & $0.9175\pm0.0006$      \\
GAT+BoT  & $0.8765\pm0.0008$ & $0.9280\pm0.0008$      \\
RevGNN-deep & $0.8774\pm0.0013$ & $0.9326\pm0.0006$      \\
RevGNN-wide & $0.8824\pm0.0015$ & $0.9450\pm0.0008$      \\
AGDN & $0.8865\pm0.0013$ & $0.9418\pm0.0005$      \\
\hline
\model-3Layer & $0.8877\pm  0.0011$ & $0.9415\pm0.0023 $      \\
\model-6Layer      & $\mathbf{0.8917\pm 0.0007}$ & $\mathbf{0.9472\pm0.0020}$      \\ \hline
\label{tab:performance}
\end{tabular}
\vspace{-0.7cm}
\end{table}
\begin{table}[!h]
\centering
\caption{Ablation study on the \textit{ogbn-proteins} dataset.}
\begin{tabular}{lcc}
\hline
Method              & Test ROC-AUC  & Validation ROC-AUC \\ \hline
\model w/o bit-wise module                 & $0.8813\pm0.0011$ & $0.9332\pm0.0009$      \\
\model w/o feature-wise module         & $0.8701\pm0.0021$ & $0.9320\pm0.0007$      \\
\model w/o edge feature           & $0.8599\pm0.0047$ & $0.9204\pm0.0038$      \\
\hline
\model         & $\mathbf{0.8917\pm 0.0007}$ & $\mathbf{0.9472\pm0.0020}$      \\ \hline
\label{tab:ab_performance}
\end{tabular}
\vspace{-0.3cm}
\end{table}


\subsection{Performance Comparison}
\tablename~\ref{tab:performance} shows the average ROC-AUC and the standard deviation for the test and validation set.
The results of the baselines are retrieved from the ogbn-proteins leaderboard\footnotemark[8]. 
Our \model outperforms all previous methods in the leaderboard and reaches an average ROC-AUC higher than 0.89 for the first time. 
Furthermore, \model only with 3 layer achieved the \textit{SOTA} performance on \textit{ogbn-proteins} dataset.
This result shows the effectiveness of our proposed bit-wise and feature-wise correlation modules, which can leverage the edge features to improve the performance by fine-grained information fusion and noise suppression. 

To further investigate the impact of each component in our proposed \model, we conduct the ablation study on the ogbn-proteins dataset. 
As shown in \tablename~\ref{tab:ab_performance}, compare with \model w/o bit-wise module, \model w/o feature-wise module. And the combination of these components (i.e., \model) yields the best performance, which indicates the necessity of bit-wise and feature-wise correlation modules.

The average training times per epoch on \textit{ogbn-proteins} with \model, 3-Layer \model, AGDN are 8.2 seconds (s), 5.9s, and 4.9s, respectively. 
The average inference times on whole graph of \textit{ogbn-proteins} with \model, 3-Layer \model, AGDN  are 11.1s, 9.3s, and 10.7s, respectively. 
Compared with ADGN, \model is slower in training speed, but has advantages in inference speed.

\subsection{Hyperparameter Analysis and Ablation Study}\label{Hyperparameter}
\begin{figure*}[t!]\label{layer}
    \centering
  \begin{subfigure}[b]{0.44\linewidth}
    \includegraphics[width=\linewidth]{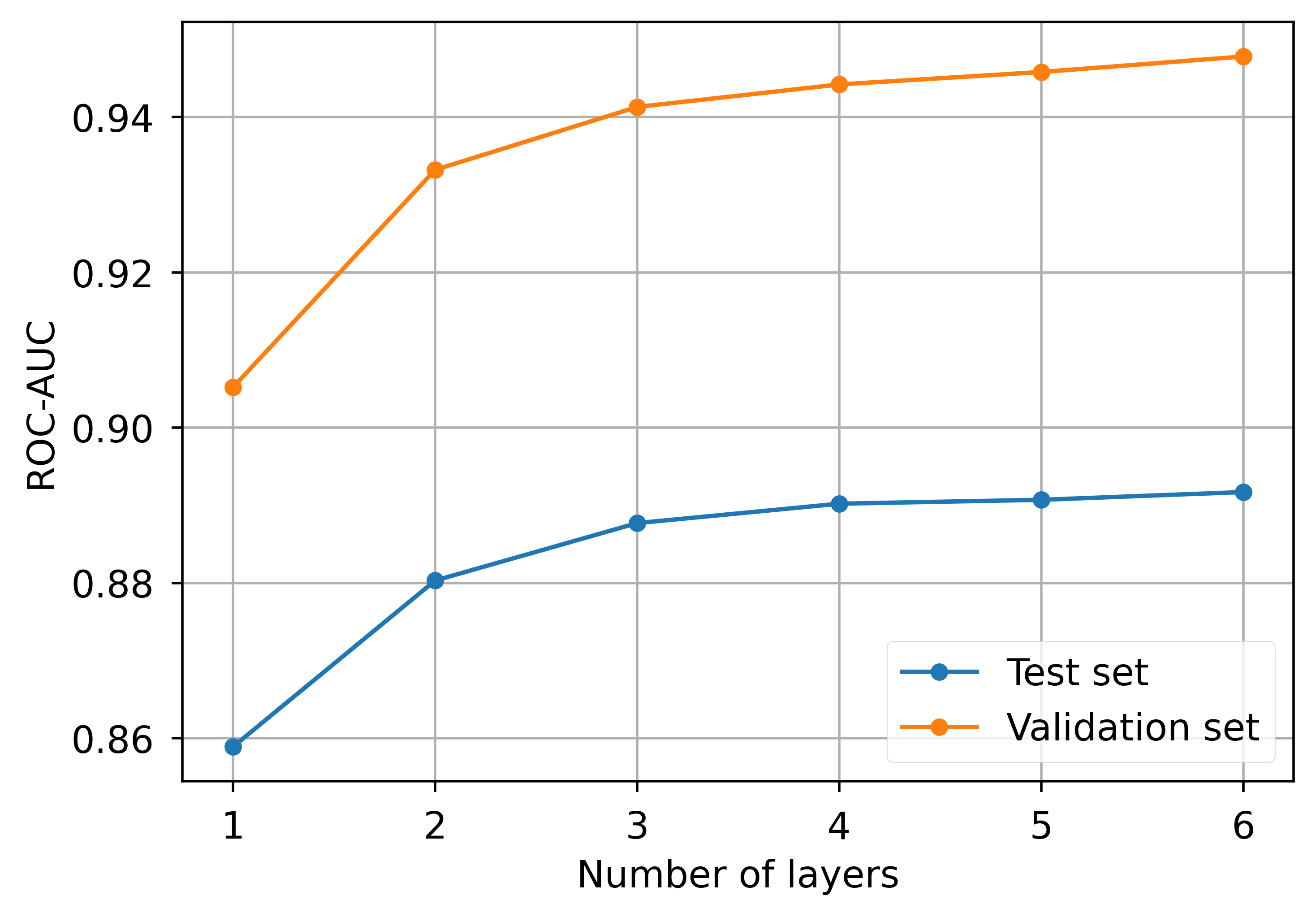}
    \caption{ }
    \label{fig:layer}
  \end{subfigure}
   \begin{subfigure}[b]{0.46\linewidth}
    \includegraphics[width=\linewidth]{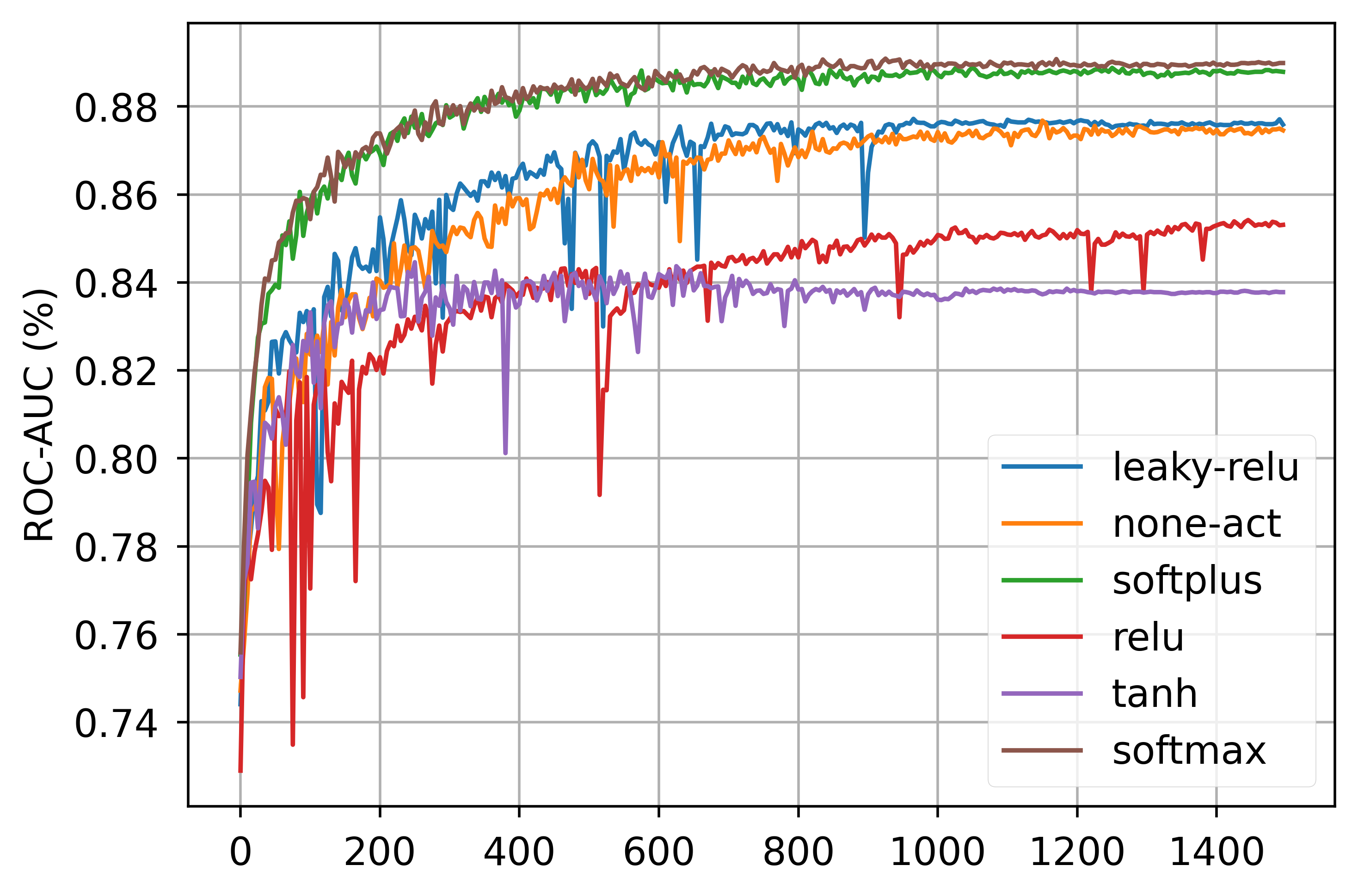}
    \caption{ }
 \label{fig:activation}
  \end{subfigure}
\vspace{-0.3cm}
\caption{Effect of hyperparameters. (a) ROC-AUC vs. the number of layers $K$. (b) Convergence on test set vs. activation function of attention.}
\vspace{-0.3cm}
\end{figure*}

\noindent \textbf{Effect of the number of layers $K$.} To study the impact of the number of layers $K$ on performance, we vary its value to $\{1, 2, 3, 4, 5 ,6\}$. As shown in Fig.~\ref{fig:layer}, with the increase of layers, the performance of the model on the test set gradually converges. However, in the comparison between the five layers and the six layers, the increase in the performance of the model on the verification set is far greater than that on the test set. This result shows that with the increase of complexity, the performance of the model can be improved, but there will be an over-fitting problem at a bottleneck. Therefore, we finally set $K$ to 6.

\noindent \textbf{Analysis on Attention Process with Correlation Module.} Because the bit-wise correlation module and the feature-wise correlation module are the most important parts of \model, we make a detailed analysis of their architecture, including activation function and edge feature.
\\
1) Activation function: as shown in Fig.~\ref{fig:activation}, by comparing these activation functions: the performance of \textit{tanh} and \textit{relu} are even worse than the one without \textit{activator}.
The model using \textit{softmax} as the activation function achieves the best performance, but the performance of model using \textit{softplus} is close to it.
This phenomenon indicates that the attention module of \model can adaptively learn the correlation weight between nodes. 
Cause $\sum_{j\in N(i)}a_{ij}$ is needed to be calculated firstly by aggregation from all neighbors, the normalization operation of \textit{softmax} is a costly operation in GNN. 
On the other hand, \model with \textit{softplus} trains faster than that with \textit{softmax} on \textit{ogbn-proteins}.
Considering the trade-off between cost and performance, \textit{softplus} is adopted on billion-scale \textit{Alipay} dataset.
\\
2) Edge feature: As shown in \tablename~\ref{tab:ab_performance}, we find that the performance of \model is better than that of \model w/o edge feature, which means that the edge feature contains the correlation information between two nodes. Therefore, it is an indispensable feature in attention processing and can help the two correlation modules get more realistic attention weights.

\begin{table}[t]
\centering
\caption{Ablation on \textit{Alipay} dataset. `S' and `D' are  primary and neighbor nodes.}
\begin{tabular}{ccccc}
\hline
$\mathcal{F}_{att}$ input & $\mathcal{F}_{prop}$ input& Feature-wise module & ROC-AUC & F1  \\ 
\hline
Edge & D. Node  & None & 0.8784 & 0.1399      \\
Edge \&  D. Node  & Edge \&  D. Node & None & 0.8916 & 0.1571    \\
Edge \&  D.+ S. Nodes & Edge \&  D. Node & None &0.8943 &0.1611    \\
Edge \&  D.+ S. Nodes & Edge \&  D. Node & Yes & 0.8961 &0.1623    \\
\hline
\label{tab:ali_performance}
\end{tabular}
\vspace{-0.6cm}
\end{table}

\section{Conclusion}
We have presented \model, a new graph attention network architecture for graph data learning.
\model consists of a bit-wise correlation module and a feature-wise correlation module, to leverage edge information and realize the fine granularity information propagation and noise filtering.
Performance evaluation on the ogbn-proteins dataset has shown that our method outperforms the state-of-the-art methods listed in the ogbn-proteins leaderboard.
And it has been tested on the billion-scale industrial dataset.
\bibliographystyle{splncs04}
\bibliography{mybibliography}
%




\end{document}